\documentclass[final]{cvpr}

\usepackage{graphicx}
\usepackage{times}
\usepackage{epsfig}
\usepackage{amsmath}
\usepackage{amssymb}
\DeclareMathOperator*{\argmin}{argmin}

\usepackage{nopageno}

\usepackage[pagebackref=true,breaklinks=true,colorlinks,bookmarks=false]{hyperref}

\def\cvprPaperID{8832} 
\def\confYear{CVPR 2021}

\begin{document}

\title{Graph Stacked Hourglass Networks for 3D Human Pose Estimation}

\author{Tianhan Xu, \quad Wataru Takano\\
Osaka University\\
{\tt\small xth430@gmail.com, takano@sigmath.es.osaka-u.ac.jp}
}

\maketitle

\begin{abstract}

   In this paper, we propose a novel graph convolutional network architecture, Graph Stacked Hourglass Networks, for 2D-to-3D human pose estimation tasks. The proposed architecture consists of repeated encoder-decoder, in which graph-structured features are processed across three different scales of human skeletal representations. This multi-scale architecture enables the model to learn both local and global feature representations, which are critical for 3D human pose estimation. We also introduce a multi-level feature learning approach using different-depth intermediate features and show the performance improvements that result from exploiting multi-scale, multi-level feature representations. Extensive experiments are conducted to validate our approach, and the results show that our model outperforms the state-of-the-art.

\end{abstract}

\begin{figure*}
\begin{center}
\includegraphics[width=\linewidth]{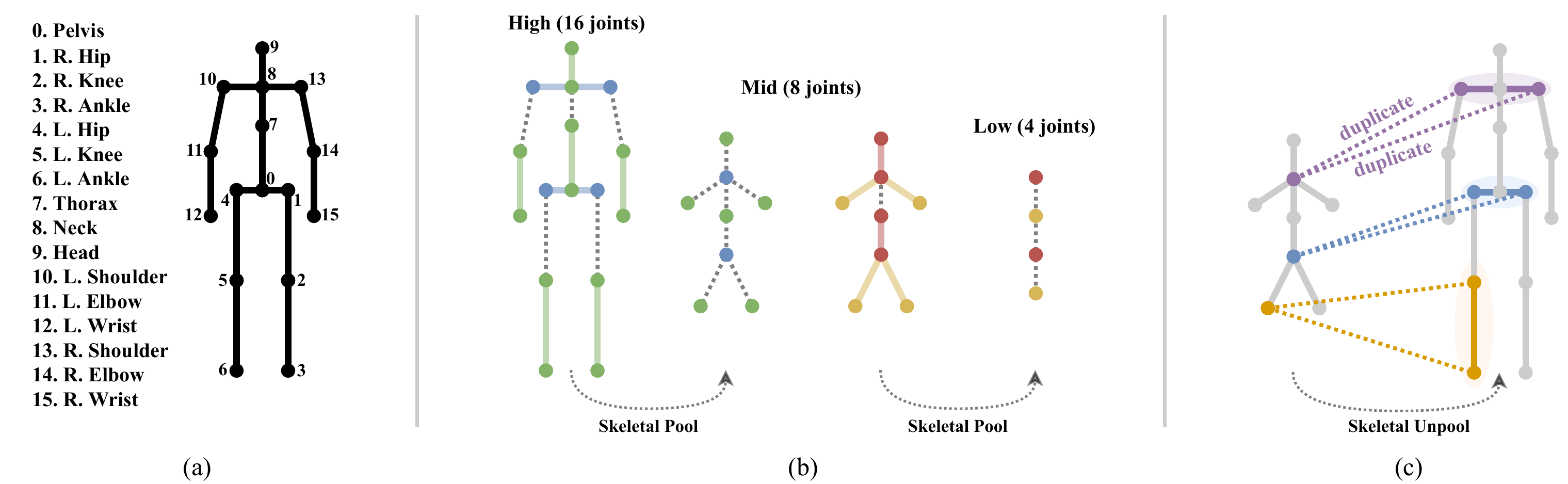}
\end{center}
   \caption{Description of the skeletal structure of the human body. 
   (a) The human skeletal graph structure consisting of 16 joints used in this paper.
   (b) Illustration of Skeletal Pool. A pair of joints composed of the same color, mapped by the max pooling operation to corresponding joints in lower scale skeletal structure.
   Here we define three different scales of skeletal structures, containing 16, 8, and 4 joints, respectively.
   (c) Illustration of Skeletal Unpool. When reverting to a higher scale skeletal structure, we duplicate the features of the lower scale joints and assign them to the two corresponding joints in the higher scale skeletal structure.}
\label{fig:skeleton}
\end{figure*}

\section{Introduction}

In recent years, with the application of deep learning methods, the performance of 2D human pose estimation has been greatly improved. Recent works show that using such detected 2D joints positions, the 3D human pose can also be efficiently and accurately regressed~\cite{Martinez2017ASY}. Due to the graph structure formed by the topology of the human skeleton, many attempts have been made to use the generalized form of CNN: Graph Convolutional Networks (GCN), to perform the regression task of 2D-to-3D human pose estimation~\cite{zhaoCVPR19semantic,wang2019gcn,Liu2020}. Since the graph convolution has a good feature extraction capability for the graph-structured data, the GCN-based approaches work well and some of them achieve the state-of-the-art results in the 2D-to-3D human pose estimation task.

However, the existing GCN-based approaches have the following limitations: 
First, the graph convolutions exploit all node information, which can be seen as that all features are processed at only ``one scale''.
Thus it is difficult to extract features that can represent spatial local and global information and limits the representation capabilities. 
Second, most existing approaches use a straightforward architecture of sequentially connecting the graph convolution layers (Fig.~\ref{fig:arch_diff}~(Top)). Such a model architecture does not take advantage of the benefits of model depth, 
such as intermediate features at each depth, and therefore limits its performance.

The core issue mentioned above is that the features extracted by existing architecture are oversimple, which limits the expressiveness of the model. 
Features with greater representation capabilities, such as multi-scale and multi-level features, are commonly used in image-related tasks.
For \textit{multi-scale} features, they denote the information from small to large resolutions of the image features, thus bringing rich image understanding from local to global~\cite{ms-cnn, FPN2017}, 
while \textit{multi-level} features denote the latent representations in different depths of the latent space, bringing important semantic information at all levels from shallow to deep.
The introduction of the above multi-scale and multi-level features can enrich the performance of the model. However, because the upsampling and downsampling operations required for multi-scale features are defined on the image, 
and the graph has an irregular structure, such methods cannot be directly applied to the graph-structured data. 

To address these issues, we propose a novel architecture for 2D-to-3D human pose estimation: \textit{Graph Stacked Hourglass Networks}.
Specifically, we do not focus on specific graph convolution operations, but rather consider how to integrate them in the architecture which gives the best performance improvement. 
Given the advantages of the 2D human pose estimation approaches, proposed architecture adopts the repeated encoder-decoder applicable to the graph-structured data for multi-scale feature extraction, 
as well as the intermediate features at each depth of the model for multi-level feature extraction. 
Such multi-scale and multi-level feature information makes the model more expressive and enables the model to achieve high-precision 3D human pose estimation.

Our work makes the following contributions. 
First, we propose \textit{Graph Hourglass} modules suitable for extracting multi-scale human skeletal features, which includes novel pooling and unpooling operations considering human skeletal structure, called \textit{Skeletal Pool} and \textit{Skeletal Unpool}.
Second, we introduce \textit{Graph Stacked Hourglass Networks (GraphSH)} consisting of the proposed \textit{Graph Hourglass} module , which incorporates multi-level feature representations at various depths of the architecture.
Our architecture incorporates multi-scale, multi-level features and a priori knowledge of the human skeleton, achieving impressive performance improvement for 2D-to-3D human pose estimation tasks.

\begin{figure*}
\begin{center}
\includegraphics[width=\linewidth]{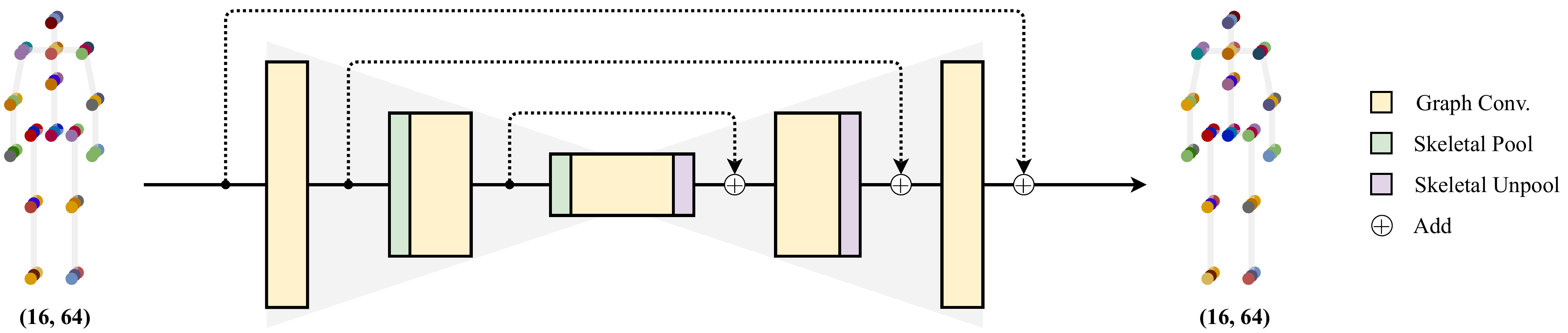}
\end{center}
   \caption{Illustration of the Graph Hourglass Module. The skeletal scale is reduced by graph convolution and skeletal pooling, up-sampled by skeletal unpooling after reaching the lowest scale, 
   and finally restored to the original skeletal scale. As the skeletal scale decrease, we increase the number of channels ($64 \rightarrow 96 \rightarrow 128$ in our experiments).
   Each graph convolution layer is followed by Batch Normalization~\cite{bn2015} and ReLU activation~\cite{relu2010}.
   Residual connections are used between features at each scale.
   Note that the inputs and outputs of the hourglass module maintain the same shape.}
\label{fig:hourglass}
\end{figure*}

\section{Related Work}\label{sec:related_work}

\textbf{3D Human Pose Estimation.}
Predicting the 3D human pose from images or videos has been an essential topic in computer vision for a long time.
In the early days, handcrafted features, perspective relationships, and geometric constraints were used to predict 3D human pose \cite{TAKANO2015116, ohashi18iros, 1542030, 6126500}.
In recent years, with the development of deep learning, there has been an increase in using deep neural networks for image-to-3D human pose estimation \cite{BMVC2016_130, tekin2017learning, pavlakos17volumetric, sun2017, yang2018, AAAI18_fang_3dpose}. 

Some methods regress 3D human pose directly from images.
Tekin~\etal~\cite{BMVC2016_130} propose a method that first train an autoencoder to learn the latent representation of the 3D human pose, then use CNN to regress the image with the latent representation, and finally connect the trained CNN with the decoder to achieve the prediction from image to 3D human pose.
Pavlakos~\etal~\cite{pavlakos17volumetric} exploit voxel to discretize representations of the space around the human body and use 3D heatmaps to estimate 3D human pose.

There are also methods that break the problem down into two steps: 
first predicting 2D human joints from the image, and then using the 2D joints information to predict 3D human pose.
Our approach falls into this category.
Martinez~\etal~\cite{Martinez2017ASY} propose a simple yet effective baseline for 3D human pose estimation that uses only 2D joints information but get highly accurate results, showing the importance of 2D joints information for 3D human pose estimation.
Since the human skeleton's topology can be viewed as a graph structure, there has been increasing use of Graph Convolutional Networks (GCN) for 2D-to-3D human pose estimation tasks~\cite{zhaoCVPR19semantic,wang2019gcn,Liu2020}.

\textbf{Graph Convolutional Networks.}
Graph Convolutional Networks (GCN) are used to perform convolution operations on graph-structured data, such as human skeleton, thus enabling effective feature extraction.
The early simple GCN is the `vanilla' GCN proposed by Kipf and Welling~\cite{Kipf:2016tc},
which consists of a simple graph convolution operation that performs the transformation and aggregation of graph-structured data, and it becomes the basic model for various graph convolution later on.
The following GCNs are based on this model with some improvements and are applied to 2D-to-3D human pose estimation.
Zhao~\etal~\cite{zhaoCVPR19semantic} propose Semantic Graph Convolution (SemGConv), which has learnable adjacency matrix parameters, 
enabling the model to learn the semantic relationships between the human joints.
In the two graph convolutions just introduced, each node information is transformed using the same weight matrix and then aggregated.
Liu~\etal~\cite{Liu2020} point out that sharing the same weight by all nodes limits the representation capabilities of graph convolution, 
and propose a new method that, first transforming each node information using different weights and then aggregating them together, called Pre-Aggregation Graph Convolution (PreAggr).
They also introduce an approach that decouples the self-connections in the graph and use separate weight to compute the self-information transformation.

The GCN using the above graph convolution for 3D human pose estimation, however, only use a straightforward overall architecture, as in Fig.~\ref{fig:arch_diff}~(Top).
Such a simple architecture prevents the model from using the multi-scale, multi-level features common in image-based tasks, limiting the performance of the model.

\textbf{Multi-scale and Multi-level Learning.}
Multi-scale and multi-level features learning drives advances in a wide variety of image-based tasks~\cite{RFB15a, DBLP:conf/eccv/LiuAESRFB16}.
Multi-scale feature learning refers to integrating features in different resolutions to provide a better understanding within the spatial domain.
Feature Pyramid Network (FPN)~\cite{FPN2017} is a powerful multi-scale feature extractor and achieves encouraging results in object detection tasks.
In the task of 2D human pose estimation from image, Hourglass structure for extracting multi-scale features allows the model to learn both local and global features, which are essential for human pose understanding.
 (\eg, spatial configuration relationships between human joints)~\cite{Newell2016StackedHN,NIPS2017_8edd7215,sun2019deep}.
Multi-level feature learning, on the other hand, represents the use of features at various depths of the network.
Some methods use a skip layer to incorporate features from the intermediate layer of the network and then combine them into the output layer~\cite{DeepEdge2015,FCN2015}.
Zhao~\etal~\cite{M2Det2019aaai} incorporate multi-scale and multi-level features, combining image pyramids of different depths to extract higher feature representations.

These image-based methods take advantage of the fact that images can be easily scaled up and down and the richness of intermediate features, thus enabling multi-scale and multi-level feature extraction.
However, due to the graph structure's irregularity, it is not trivial to scale it up or down like an image, so such approaches have not been applied much to tasks with graph-structured data.

In the next section, we present our proposed novel graph convolutional network architecture that integrates multi-scale and multi-level features of the graph-structured data.

\section{Graph Stacked Hourglass Networks}

\begin{figure*}
\begin{center}
\includegraphics[width=\linewidth]{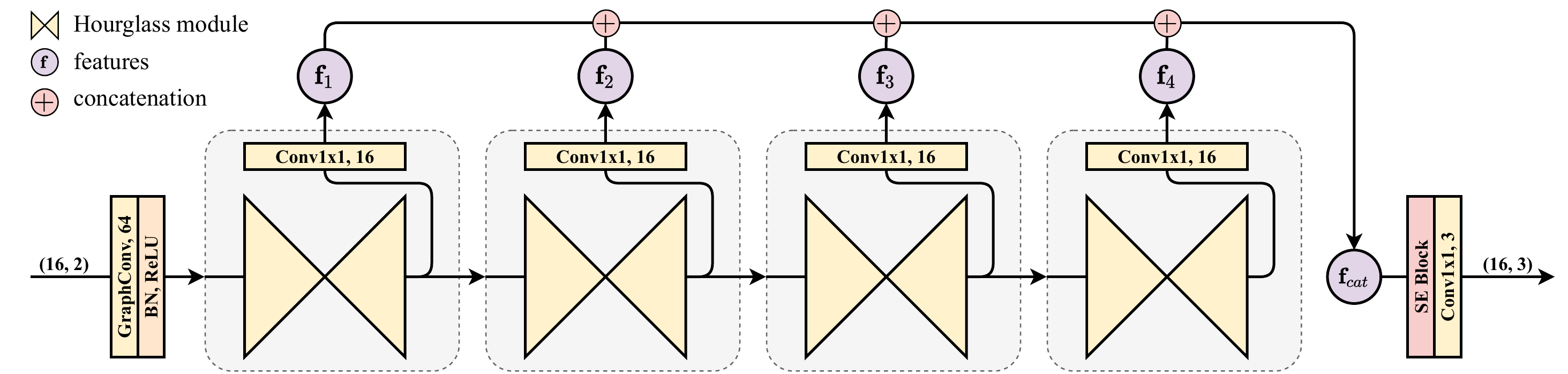}
\end{center}
   \caption{The overall architecture of our network. The input 2D joints are fed into the hourglass module after the pre-processed graph convolution layer, 
   and the outputs of the hourglass module are both processed as intermediate features, and also fed into the subsequent hourglass modules, except the last one. 
   All the intermediate features are concatenated and entered into the SE block~\cite{senet2018}, then the final feature is passed through a 1x1 convolutional layer to output the final 3D pose prediction.
   The number of the convolution module indicates the number of channels of the output. Our network takes a 4-stacking approach. The input and output of the hourglass module keep the dimension of 64 channels.
   Note that the graph structure is maintained throughout the whole network.}
\label{fig:sh_network}
\end{figure*}

\subsection{Hourglass Module}
Our approach is inspired by Stacked Hourglass Networks proposed by Newell~\etal~\cite{Newell2016StackedHN} for estimating 2D human pose from images,
which exploits repeated hourglass-like encoder-decoder architecture.
We aim to extend such an hourglass structure to the graph for extracting multi-scale features of the graph-structured data.

Using deep neural networks for computer vision tasks, multi-scale features of the image are essential for image understanding.
Since images have large amounts of information at high resolution, the model can extract much detailed information from them. Alternatively, while the image is at low resolution, the model can better extract globalized information. 
The hourglass structure, accompanied by downsampling and upsampling operations, enables the image features to go through all resolutions so that crucial information can be extracted at all scales.
Previous works have shown that this hourglass structure has strong feature extraction capabilities, especially for tasks requiring both local and global information, such as human pose estimation~\cite{Newell2016StackedHN, NIPS2017_8edd7215}.
Moreover, stacking such a structure enables repeated feature extraction and enhances model performance.

Our motivation is to extend such a structure with powerful multi-scale feature extraction capabilities to graph-structured data to achieve highly accurate 2D-to-3D human pose estimation.
In the hourglass structure, downsampling and upsampling of the data are implemented by pooling and unpooling operations, respectively. 
Such operations are easy to define on the images because of their regularized structure, but there is no consistent way to define them on the graph.
On this point, there are some previous works regarding pooling and unpooling operations on the graph~\cite{gao2019graph, Lee2019SelfAttentionGP}. 
However, these pooling and unpooling operations are defined on the more generalized arbitrary-shaped graph structure, while the human skeletal graph used in this study has a fixed structure as shown in Fig.~\ref{fig:skeleton}(a). 
Here, we exploit this property to propose pooling and unpooling methods applicable to the human skeletal graph, called \textit{Skeletal Pooling} and \textit{Skeletal Unpooling}.

\textbf{Skeletal pooling.}
According to the property of the human body structure, we group the human body nodes in pairs, where the corresponding two nodes' features are fused into one node in the lower-scale skeleton structure, using max pooling operation.
With repeated pooling operation, we can obtain three different scales of skeletal structures containing 16, 8, and 4 nodes, respectively, and each of them corresponds to a different graph structure.
For relatively lower-scale graph representations, we use more channels to encode information to prevent information degradation.
The illustration of skeletal pooling and the three-scales skeleton graph structures are shown in Fig.~\ref{fig:skeleton}(b).

\textbf{Skeletal unpooling.}
We use unpooling operation to restore lower-scale skeletal structures to their original size, enabling them to fuse higher-scale skeletal information and pass it to subsequent processing.
We adopt a very simple approach: since lower-scale nodes are generated by two grouped higher-scale nodes, in unpooling operation, we duplicate the feature representations of the lower-scale node and assign them to corresponding two nodes to recover the higher-scale skeletal representations.
The illustration of skeletal unpooling is shown in Fig.~\ref{fig:skeleton}(c).

\textbf{Graph hourglass design.}
Using the above skeletal pooling and unpooling operations, we propose a novel graph hourglass module applicable to human skeleton representation.The detailed hourglass structure is shown in Fig.~\ref{fig:hourglass}.
For the same scale of the skeletal structure, we applied the residual connections \cite{He2016DeepRL} to pass information and prevent the vanishing gradient problem.
Note that our hourglass structure does \textit{not depend on} the specific graph convolution layer, so arbitrary graph convolution operation can be implemented on our model, such as the three introduced in Sect.~\ref{sec:related_work}.

Graph U-Nets proposed by Gao~\etal~\cite{gao2019graph} is the closest work to our architecture, but differs in two ways.
First, \cite{gao2019graph} uses an input-related dynamic pooling operation so that different pooled skeletal structures are obtained depending on the input.
Our method utilizes a priori knowledge of human skeletal structure, and this approach is more suitable for feature extraction of specific skeletal structures while ensuring the stability of pooling.
Second, we use more channels at relatively low scales of skeleton representation to reduce information loss due to scale changes and make our architecture more expressive, while Graph U-Nets uses the same number of channels at all scales.

\subsection{Multi-scale and Multi-level Features}\label{sec:msml}
Features are extracted at multi-scales as the hourglass module processes the information across three skeletal structures.
As multi-scale features that can represent information on the spatial aspect of the graph, we believe that multi-level features in terms of the depth of latent space can also bring valuable information to the final prediction.
Specifically, we integrate the intermediate features at each depth level of the network for the final 3D human pose estimation.
As shown in Fig.~\ref{fig:sh_network}, we use the spatial 1x1 convolution to reduce the channels of intermediate features and concatenate them into an overall feature representation $\mathbf{f}_{cat}$.
For the architecture that stacks $n$ hourglass modules, the overall feature can be represented as:
\begin{equation}
\label{eq:multi_level_feature}
   \mathbf{f}_{cat} = \mathrm{Concat}(\mathbf{f}_1, \mathbf{f}_2, \cdots, \mathbf{f}_n) \in \mathbb{R}^{K \times C}, ~ \mathbf{f}_i \in \mathbb{R}^{K \times \frac{C}{n}} .
\end{equation}
Here $K$ is the number of joints, and $C$ represents the number of input and output channels of the hourglass module ($C=64$ in our experiments).
Then, we follow the Squeeze-and-Excitation block (SE block)~\cite{senet2018} to enhance important semantic information in multi-level features.
SE block computes the channel-wise weights of the overall feature, and since the overall feature is concatenated by multi-level features along the channel axis, 
this block enables the model to extract the more semantically meaningful feature representations among each intermediate feature.
Specifically, first we use global average pooling to transform the overall feature $\mathbf{f}_{cat} \in \mathbb{R}^{K \times C}$ into a channel-wise statistics $\mathbf{z} \in \mathbb{R}^C$; then we use $\mathbf{z}$ to calculate the channel-wise dependency $\mathbf{s}$ as follows:
\begin{equation}
\label{eq:ae_weight}
   \mathbf{s} = \mathrm{Sigmoid}(\mathbf{W}_2 ~ \mathrm{ReLU}(\mathbf{W}_1 \mathbf{z})) .
\end{equation}
Here $\mathbf{W_1} \in \mathbb{R}^{\frac{C}{r} \times C}$, $\mathbf{W_2} \in \mathbb{R}^{ C \times \frac{C}{r}}$, and $r$ indicates reduction ratio ($r=8$ in our experiments).
Finally, the features of each channel are weighted by $\mathbf{s}$ as follows.
\begin{equation}
\label{eq:ae_fin}
   \tilde{\mathbf{f}}_{cat} = \mathbf{f}_{cat} \odot \mathbf{s} ,
\end{equation}

The output feature of SE block $\tilde{\mathbf{f}}_{cat}$ is then fed to the output layer for final prediction.

\subsection{Network Architecture}

As in previous works~\cite{Newell2016StackedHN, NIPS2017_8edd7215}, our backbone network consists of the proposed graph hourglass module stacked.

The input 2D joint information is first mapped to the latent feature space via a pre-processing graph convolution layer.
The features that through the hourglass module are fed into the 1x1 convolution layer to be transformed into intermediate features, and also passed to the next hourglass module, except the last one.
All the intermediate features are concatenated into the final feature, and the SE block adjusts its channel-wise weights, 
and which is then fed into the output convolution layer and mapped to the output space.
Our overall network structure is shown in Fig.~\ref{fig:sh_network}.

In the following experiments, our model uses PreAggr~\cite{Liu2020} as the graph convolution layer, with 4-stacking hourglass approaches and latent space of 64 channels.

\begin{figure}
\begin{center}
\includegraphics[width=\linewidth]{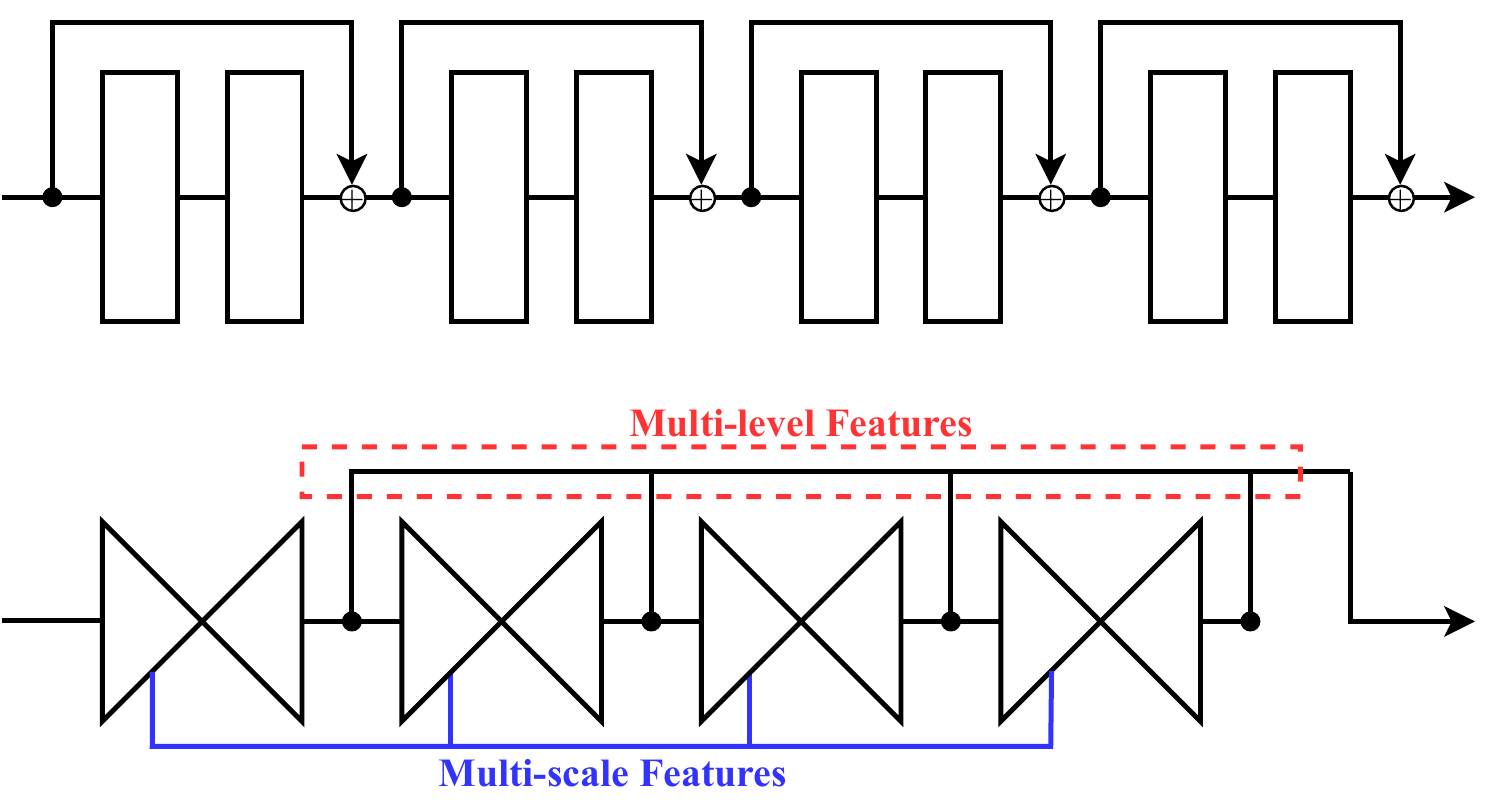}
\end{center}
   \caption{Comparison of the model architecture.
   Top: Most used Sequential Residual blocks architecture (SeqRes) for GCN-based 2D-to-3D human pose estimation \cite{Liu2020,zhaoCVPR19semantic,wang2019gcn}. The rectangles represent the graph convolution layers.
   Bottom: Proposed Graph Stacked Hourglass Networks architecture. Multi-scale, multi-level features are indicated in the figure. Refer to Fig.~\ref{fig:sh_network} for detailed architecture.
   }
\label{fig:arch_diff}
\end{figure}

\section{Experiments}
In this section, we first describe the experimental setup for 2D-to-3D human pose estimation tasks.
Next, we introduce the dataset used and its evaluation protocols.
Then several ablation studies are conducted regarding the proposed architecture.
Finally, we show our experimental results and comparisons with state-of-the-art methods.

\subsection{2D-to-3D human pose estimation}
Our goal is to predict 3D joint positions in the camera coordinate system with given 2D joint positions in the pixel coordinate system.
Specifically, 2D joints consisting of $K$ nodes are described by $\mathbf{x} \in \mathbb{R}^{K\times 2}$, the corresponding 3D joints are $\mathbf{y} \in \mathbb{R}^{K\times 3}$.
The model aims to learn a mapping $f^*: \mathbb{R}^{K\times 2} \rightarrow \mathbb{R}^{K\times 3}$ that minimizes the errors over a dataset containing $N$ poses.

\begin{equation}
\label{eq:loss}
   f^* = \argmin_f\frac{1}{N}\sum_{i = 1}^{N}\mathcal{L} \left(f(\mathbf{x}_i), \mathbf{y}_i \right).
\end{equation}

In this study, Mean Squared Error (MSE) is used as a loss function $\mathcal{L}$ for training.

\subsection{Datasets and Evaluation Protocols}

\textbf{Datasets.} 
The Human3.6M dataset~\cite{ionescu2013human3} is the most widely used dataset in the 3D human pose estimation tasks. It uses motion captures to obtain the 3D pose information of the subjects and 4 cameras with different orientations to record the corresponding video image information. 
The provided camera parameters allow us to obtain the ground truth of the corresponding 2D joint coordinates in each image frame.
The dataset provides 3.6 million images by recording 11 professional actors performing 15 different actions, such as eating, walking, etc.
In the following experiments, we mainly use the Human3.6M for training and testing.
The MPI-INF-3DHP test set~\cite{mono-3dhp2017} provides images in three different scenarios: studio with a green screen (GS), studio without green screen (noGS) and outdoor scene (Outdoor).
We use this dataset to test the generalization capabilities of our proposed architecture.

\textbf{Evaluation protocols.}
For the Human3.6M, There are two evaluation protocols used in previous works~\cite{Martinez2017ASY,zhaoCVPR19semantic,Liu2020,wang2019gcn}.
\textit{Protocol \#1} uses the Mean Per Joint Position Error (MPJPE) in millimeter as evaluation metric, which calculates the Euclidean distance error between the prediction and the ground truth after the origin (pelvis) alignment.
\textit{Protocol \#2} aligns the prediction with the ground truth by rigid transformation and then calculates the error.
In this study, we use \textit{Protocol \#1} to evaluate our approach since performance under both protocols is usually consistent and \textit{Protocol \#1} is more appropriate for our experimental setup.
For the MPI-INF-3DHP test set, we follow previous works~\cite{luo2018orinet, wang2019gcn} and use 3D-PCK and AUC as evaluation metrics.

\subsection{Implementation Details}
Our implementation follows the settings of previous works~\cite{Martinez2017ASY,zhaoCVPR19semantic,Liu2020}.
As introduced in \cite{pavllo:videopose3d:2019}, we normalize the coordinates of the 2d and 3d joints and align the root joint (pelvis) to the origin.

In our experiments, we use a 4-stacked hourglass architecture. Previous works typically use 128 as the number of channels~\cite{zhaoCVPR19semantic, wang2019gcn, Liu2020}, 
but due to the relative complexity of our model architecture, we use 64 channels to keep the number of parameters at the same scale as previous works.
We use Adam~\cite{kingma:adam} as the optimizer with an initial learning rate set to 0.0001 and decay by 0.92 per 20,000 iterations. We use a mini-batch size of 256.
Since the learning of graph-structured data is very prone to overfitting, we apply Dropout~\cite{JMLR:v15:srivastava14a} with a dropout probability of 0.25 to all graph convolutional layers within the hourglass module.
The entire training follows an end-to-end fashion.

\begin{table}
\begin{center}
\begin{tabular}{l|c|c}
\hline
Method & params & MPJPE (mm)\\
\hline\hline
gPool/gUnpool~\cite{gao2019graph} & 3.70M & 40.6 \\
SAGPool~\cite{Lee2019SelfAttentionGP}/gUnpool~\cite{gao2019graph} & 3.70M & 41.5 \\
w/o Pool/Unpool & 6.86M & 38.2 \\
\hline
Skeletal Pool/Unpool & 3.70M & \textbf{35.8}\\
\hline
\end{tabular}
\end{center}
\caption{Ablation study of pool/unpool operation. Various methods of graph pool/unpool are used, and the method that remove the pool/unpool layer is also added for comparison.}
\label{tbl:ablation_pool}
\end{table}

\begin{table}
\begin{center}
\begin{tabular}{l|c|c}
\hline
Method & params & MPJPE (mm)\\
\hline\hline
No ML-F & 3.70M & 38.0 \\
ML-F w/o SE block & 3.70M & 37.1 \\
ML-F w/ SE block (Ours) & 3.70M & \textbf{35.8} \\
\hline
\end{tabular}
\end{center}
\caption{Ablation study of multi-level features (ML-F). With almost the same number of parameters, our proposed approach using multi-level features with SE block achieves the best results.}
\label{tbl:ablation_ml}
\end{table}

\begin{table}
\begin{center}
\begin{tabular}{ll|c|c}
\hline
Method &  & params & MPJPE (mm)\\
\hline\hline
Vanilla~\cite{Kipf:2016tc} & SeqRes & 0.14M & 94.4\\
   & GraphSH & 0.22M & \textbf{59.1}\\
\hline
SemGConv~\cite{zhaoCVPR19semantic} & SeqRes & 0.27M & 52.5\\
   & GraphSH & 0.44M & \textbf{39.2}\\
\hline
PreAggr~\cite{Liu2020} & SeqRes & 4.22M & 37.8\\
   & GraphSH & 3.70M & \textbf{35.8}\\
\hline
\end{tabular}
\end{center}
\caption{Comparison of Sequential Residual blocks (SeqRes) (Fig.~\ref{fig:arch_diff}~(Top)) and Graph Stacked Hourglass (GraphSH) architectures on 2d-to-3d human pose estimation errors. Three different graph convolution operations are used.}
\label{tbl:ablation_hourglass}
\end{table}

\begin{table*}
\begin{center}
\resizebox{\textwidth}{!}{%
\begin{tabular}{@{}lrrrrrrrrrrrrrrrr@{}}
\hline\noalign{\smallskip}
\textbf{Protocol \#1} & Direct. & Discuss & Eating & Greet & Phone & Photo & Pose & Purch. & Sitting & SittingD. & Smoke & Wait & WalkD. & Walk & WalkT. & Avg. \\
\noalign{\smallskip}\hline\hline\noalign{\smallskip}
Lee~\etal~\cite{Lee2018PropagatingL3} ECCV'18 $(\dagger)$ & 40.2 & 49.2 & 47.8 & 52.6 & 50.1 & 75.0 & 50.2 & 43.0 & 55.8 & 73.9 & 54.1 & 55.6 & 58.2 & 43.3 & 43.3 & 52.8 \\
Pavllo~\etal~\cite{pavllo:videopose3d:2019} CVPR'19 $(\dagger)$ & 45.2 & 46.7 & 43.3 & 45.6 & 48.1 & 55.1 & 44.6 & 44.3 & 57.3 & 65.8 & 47.1 & 44.0 & 49.0 & 32.8 & 33.9 & 46.8\\
Cai~\etal~\cite{cai2019exploiting} ICCV'19 $(\dagger)$ & 44.6 & 47.4 & 45.6 & 48.8 & 50.8 & 59.0 & 47.2 & 43.9 & 57.9 & 61.9 & 49.7 & 46.6 & 51.3 & 37.1 & 39.4 & 48.8 \\
Xu~\etal~\cite{Xu_2020_CVPR} CVPR'20 $(\dagger)$ & 37.4 & 43.5 & 42.7 & 42.7 & 46.6 & 59.7 & 41.3 & 45.1 & 52.7 & 60.2 & 45.8 & 43.1 & 47.7 & 33.7 & 37.1 & 45.6 \\

\noalign{\smallskip}\hline\noalign{\smallskip}
Martinez~\etal~\cite{Martinez2017ASY} ICCV'17 & 51.8 & 56.2 & 58.1 & 59.0 & 69.5 & 78.4 & 55.2 & 58.1 & 74.0 & 94.6 & 62.3 & 59.1 & 65.1 & 49.5 & 52.4 & 62.9 \\
Tekin~\etal~\cite{tekin2017learning} ICCV'17 & 54.2 & 61.4 & 60.2 & 61.2 & 79.4 & 78.3 & 63.1 & 81.6 & 70.1 & 107.3 & 69.3 & 70.3 & 74.3 & 51.8 & 63.2 & 69.7 \\
Sun~\etal~\cite{sun2017} ICCV'17 $(+)$ & 52.8 & 54.8 & 54.2 & 54.3 & 61.8 & 67.2 & 53.1 & 53.6 & 71.7 & 86.7 & 61.5 & 53.4 & 61.6 & 47.1 & 53.4 & 59.1 \\
Yang~\etal~\cite{yang2018} CVPR'18 $(+)$ & 51.5 & 58.9 & 50.4 & 57.0 & 62.1 & 65.4 & 49.8 & 52.7 & 69.2 & 85.2 & 57.4 & 58.4 & \textbf{43.6} & 60.1 & 47.7 & 58.6\\
Fang~\etal~\cite{AAAI18_fang_3dpose} AAAI'18 & 50.1 & 54.3 & 57.0 & 57.1 & 66.6 & 73.3 & 53.4 & 55.7 & 72.8 & 88.6 & 60.3 & 57.7 & 62.7 & 47.5 & 50.6 & 60.4\\
Pavlakos~\etal~\cite{Pavlakos2018} CVPR'18 $(+)$ & 48.5 & 54.4 & 54.5 & 52.0 & 59.4 & 65.3 & 49.9 & 52.9 & 65.8 & 71.1 & 56.6 & 52.9 & 60.9 & 44.7 & 47.8 & 56.2 \\
Zhao~\etal~\cite{zhaoCVPR19semantic} CVPR'19 & 48.2 & 60.8 & 51.8 & 64.0 & 64.6 & \textbf{53.6} & 51.1 & 67.4 & 88.7 & \textbf{57.7} & 73.2 & 65.6 & 48.9 & 64.8 & 51.9 & 60.8 \\
Sharma~\etal~\cite{sharma2019} ICCV'19 & 48.6 & 54.5 & 54.2 & 55.7 & 62.2 & 72.0 & 50.5 & 54.3 & 70.0 & 78.3 & 58.1 & 55.4 & 61.4 & 45.2 & 49.7 & 58.0 \\
Ci~\etal~\cite{wang2019gcn} ICCV'19 $(+)(*)$ & 46.8 & 52.3 & \textbf{44.7} & \textbf{50.4} & \textbf{52.9} & 68.9 & 49.6 & 46.4 & 60.2 & 78.9 & \textbf{51.2} & 50.0 & 54.8 & 40.4 & \textbf{43.3} & 52.7 \\
Liu~\etal~\cite{Liu2020} ECCV'20 & 46.3 & 52.2 & 47.3 & 50.7 & 55.5 & 67.1 & 49.2 & \textbf{46.0} & 60.4 & 71.1 & 51.5 & 50.1 & 54.5 & 40.3 & 43.7 & 52.4 \\

\noalign{\smallskip}\hline\noalign{\smallskip}
Ours & \textbf{45.2} & \textbf{49.9} & 47.5 & 50.9 & 54.9 & 66.1 & \textbf{48.5} & 46.3 & \textbf{59.7} & 71.5 & 51.4 & \textbf{48.6} & 53.9 & \textbf{39.9} & 44.1 & \textbf{51.9} \\
\noalign{\smallskip}\hline
\end{tabular}}
\end{center}
\caption{Quantitative evaluation results using MPJPE in millimeter on Human3.6M~\cite{ionescu2013human3} under \textit{Protocol \#1}, no rigid alignment or transform applied in post-processing. CPN~\cite{Chen2018CPN} detections 2D keypoints are used as input. $(+)$ uses extra data from MPII~\cite{Andriluka20142DHP}. $(\dagger)$ uses temporal information. Best in bold.}
\label{tbl:h36m_p1_cpn}
\end{table*}

\begin{table*}
\begin{center}
\resizebox{\textwidth}{!}{%
\begin{tabular}{@{}lrrrrrrrrrrrrrrrr@{}}
\hline\noalign{\smallskip}
\textbf{Protocol \#1} & Direct. & Discuss & Eating & Greet & Phone & Photo & Pose & Purch. & Sitting & SittingD. & Smoke & Wait & WalkD. & Walk & WalkT. & Avg. \\
\noalign{\smallskip}\hline\hline\noalign{\smallskip}
Zhou~\etal~\cite{Zhou2019HEMletsPL} ICCV'19 $(+)$ & 34.4 & 42.4 & 36.6 & 42.1 & 38.2 & 39.8 & 34.7 & 40.2 & 45.6 & 60.8 & 39.0 & 42.6 & 42.0 & 29.8 & 31.7 & 39.9\\
Ci~\etal~\cite{wang2019gcn} ICCV'19 $(+)(*)$ & 36.3 & 38.8 & 29.7 & 37.8 & 34.6 & 42.5 & 39.8 & 32.5 & 36.2 & 39.5 & 34.4 & 38.4 & 38.2 & 31.3 & 34.2 & 36.3\\

\noalign{\smallskip}\hline\noalign{\smallskip}
Martinez~\etal~\cite{Martinez2017ASY} ICCV'17 & 45.2 & 46.7 & 43.3 & 45.6 & 48.1 & 55.1 & 44.6 & 44.3 & 57.3 & 65.8 & 47.1 & 44.0 & 49.0 & 32.8 & 33.9 & 46.8\\
Zhao~\etal~\cite{zhaoCVPR19semantic} CVPR'19 & 37.8 & 49.4 & 37.6 & 40.9 & 45.1 & \textbf{41.4} & 40.1 & 48.3 & 50.1 & \textbf{42.2} & 53.5 & 44.3 & 40.5 & 47.3 & 39.0 & 43.8\\
Liu~\etal~\cite{Liu2020} ECCV'20 & 36.8 & 40.3 & 33.0 & 36.3 & 37.5 & 45.0 & 39.7 & 34.9 & 40.3 & 47.7 & 37.4 & 38.5 & 38.6 & 29.6 & 32.0 & 37.8\\

\noalign{\smallskip}\hline\noalign{\smallskip}
Ours & \textbf{35.8} & \textbf{38.1} & \textbf{31.0} & \textbf{35.3} & \textbf{35.8} & 43.2 & \textbf{37.3} & \textbf{31.7} & \textbf{38.4} & 45.5 & \textbf{35.4} & \textbf{36.7} & \textbf{36.8} & \textbf{27.9} & \textbf{30.7} & \textbf{35.8}\\
\noalign{\smallskip}\hline
\end{tabular}}
\end{center}
\caption{Quantitative evaluation results using MPJPE in millimeter on Human3.6M~\cite{ionescu2013human3} under \textit{Protocol \#1}, no rigid alignment or transform applied in post-processing. Ground truth 2D keypoints are used as input. $(+)$ uses extra data from MPII~\cite{Andriluka20142DHP}. $(*)$ uses pose scales in both training and testing. Best in bold.}
\label{tbl:h36m_p1_gt}
\end{table*}

\subsection{Ablation Study}\label{sec:ablation}

\textbf{Pooling and Unpooling.}
Pooling and unpooling layers play an important role in the hourglass module.

Due to the graph structure's irregularity, there is no consistent way to pool graph-structured data, so we compare the impact of different pooling methods on model performance.
Here we compare the performance of three graph pooling operations: gPool~\cite{gao2019graph}, SAGPool~\cite{Lee2019SelfAttentionGP}, and our proposed Skeletal Pool.

The first two pooling methods take the same idea: calculate the scores of each node by some operation, and keep the part of the node with the higher score. 
Such pooling methods are initially designed for more general graph pooling situations, with the benefit that pooling operations can be defined for different graph-structured inputs.
However, since the human skeleton used in this study has a fixed graph structure, it does not take advantage of the benefits of these pooling approaches. 
Moreover, since these pooling methods are \textit{input-dependent}, different subgraphs are generated based on different inputs, which not only introduces computational complexity but also makes it difficult for the model to learn valuable features stably.

Compared to the above pooling methods that require computation, our pooling approach is more like methods that focus on geometric information of graph structure (\eg,~edges, nodes), such as mesh sampling~\cite{COMA:ECCV2018} or edge contraction~\cite{Surface} in mesh convolution.
Specifically, We follow the node grouping method in \cite{zhaoCVPR19semantic} to perform pooling operations on paired nodes. In their work, the features lose their graph structure after pooling.
We extend their grouping concept by designing three subgraph structures of the human body consisting of 16, 8, and 4 nodes, respectively.
Such a pooling approach not only exploits the topology of the human skeleton, which is more interpretable relative to other pooling approaches, 
but also, due to its simplicity, greatly reduces the computational complexity of the pooling layer and improves the speed of training and inference.
Moreover, the two pooling methods mentioned above completely discard some nodes' information, resulting in some degree of information loss. Instead, our proposed Skeletal Pool refers to the concept of image pooling, which performs a maximum pooling operation between two nodes, thus can summarize the information of both nodes.

Regarding unpooling, in Graph U-Net~\cite{gao2019graph}, the gUnpool operation assigns zero vectors to nodes that are not selected at the time of pooling, 
which loses much valuable information and results in more sparse features. Experimental results show that such an unpooling operation is not suitable for understanding structures with few nodes like the human skeleton.
On the contrary, our proposed Skeletal Unpool operation copies and assigns the node features in the lower-scale graph representation to the corresponding two nodes in the higher-scale graph, passes them into the following graph convolutional layer, makes our model have better representation capability.

We conduct experiments based on three pool/unpool settings, and the results are shown in Table~\ref{tbl:ablation_pool}. We also add a comparison that removes all the pool/unpool layers to validate the importance of the proposed skeletal pool/unpool approaches.
Note that we use ground truth 2D joints as input in this and the following ablation experiments to eliminate the influence of the 2D human pose detector.

\begin{figure*}
\begin{center}
\includegraphics[width=\linewidth]{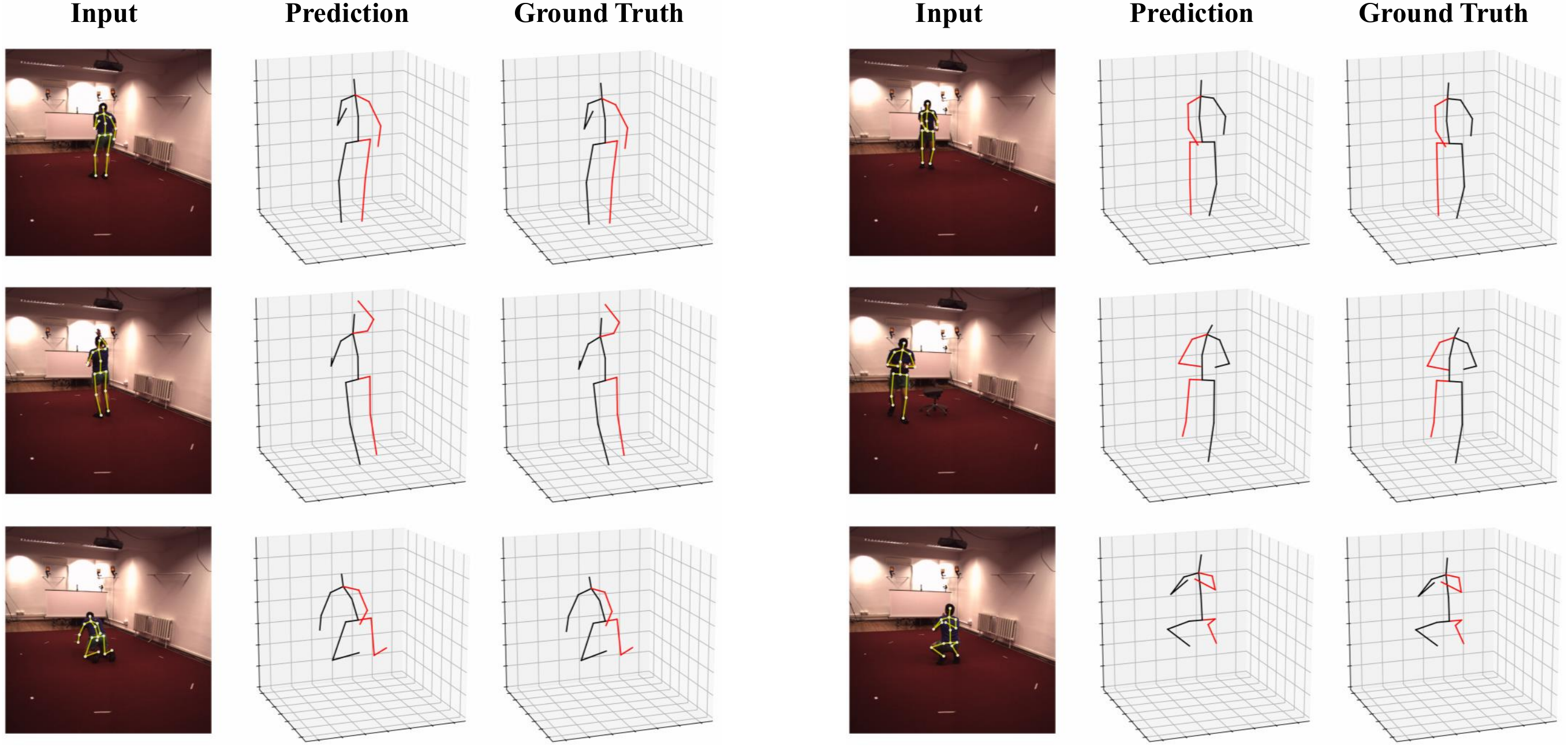}
\end{center}
   \caption{Qualitative results of our method on Human3.6M~\cite{ionescu2013human3}.}
\label{fig:viz_result}
\end{figure*}

\textbf{Multi-scale and Multi-level Features.}
Multi-scale feature extraction is achieved by pooling and unpooling in the hourglass module that transforms features across three different scales. 
For comparison, we remove all pooling and unpooling layers in our architecture, which means that the features are always processed at the highest scale.
The results in Table~\ref{tbl:ablation_pool} show that multi-scale features derived from skeletal pooling and unpooling can improve model performance.
For multi-level features, our architecture concatenates each level's intermediate features and feeds them into the SE block to obtain the final overall feature.
We compare three different settings:
(1)~No multi-level features are used. At this point the model simply connect all the hourglass modules sequentially, and the last hourglass modules is connected to the final output layer of 1x1 convolution.
(2)~Remove the SE block that calculates the weights of each intermediate feature.
(3)~Our proposed GraphSH architecture.
Results are shown in Table~\ref{tbl:ablation_ml}.

\textbf{Stacked Hourglass Architecture.}
To verify that our architecture has better performance than the simple Sequential Residual blocks (denoted as \textit{SeqRes}) in previous works \cite{zhaoCVPR19semantic,Liu2020,wang2019gcn}, 
we use Vanilla Graph Convolution, Semantic Graph Convolution (SemGConv), and Pre-aggregation Graph Convolution (PreAggr) introduced in Sect.~\ref{sec:related_work} as convolution layers in our GraphSH architecture, respectively, and compare them to the corresponding SeqRes models.
To make the comparison fair, we reduce the number of channels in the convolution to 64, so that the overall number of parameters is at the same scale as the SeqRes model.
Results are shown in Table~\ref{tbl:ablation_hourglass}.

The results show that the model using our architecture performs better even though we use fewer channels than other GCN-based approaches. Our architecture does not rely on specific graph convolution layers, 
which suggests that any graph convolution layers for 3D human pose estimation can be applied to our architecture and improve its performance compared to the SeqRes architecture. 

Moreover, our hourglass module can be seen as a well-integrated, high-performance graph convolution module, indicating that our hourglass module is general and can be easily extended to other tasks using graph convolution, such as action recognition~\cite{stgcn2018aaai, Li_2019_CVPR}, motion prediction~\cite{Li2020DenselyCG}, etc.

\subsection{Comparison with the State-of-the-Art}
We use two types of 2D joint detection data for evaluation: Cascaded Pyramid Network (CPN)~\cite{Chen2018CPN} detections and ground truth 2D keypoints, the results are shown in Table~\ref{tbl:h36m_p1_cpn}~and Table~\ref{tbl:h36m_p1_gt}, respectively.
Among the other methods, some use temporal information~\cite{Lee2018PropagatingL3,pavllo:videopose3d:2019,cai2019exploiting,Xu_2020_CVPR}, some use additional data for training~\cite{sun2017,yang2018,Pavlakos2018}, and some use 3D pose scale in both training and testing~\cite{wang2019gcn}. 
These results suggest that our approach outperforms the state-of-the-art.

Table~\ref{tbl:h36m_p1_gt} show that when given precise 2D joint information, the performance improvement of our model is significant, outperforming other GCN-based methods by a large margin.
Therefore, we believe that in combination with methods that refine detected 2D joint information, such as deep kinematic analysis~\cite{Xu_2020_CVPR}, the performance of our method on noisy 2D joints can be enhanced even more.
Compared to the second-place method~\cite{Liu2020} (4.38M), our model uses fewer parameters (3.70M), showing that our architecture can extract more important features, which provides a better understanding of the human pose.

To evaluate the generalization capabilities of our approach to domain shift, we apply the model trained on the Human3.6M to the MPI-INF-3DHP test set. Results are shown in Table~\ref{tbl:mpi-inf-3dhp}.
Although we train the model using only the Human3.6M, our approach outperforms the others, indicating that our architecture has strong generalization capabilities to unseen datasets.

The qualitative results of our method are shown in Fig.~\ref{fig:viz_result}.

\begin{table}
\begin{center}
\resizebox{\columnwidth}{!}%
{
\begin{tabular}{l|c|ccccc}
\hline
   & Training data & GS & noGS & Outdoor
   & \begin{tabular}{c}
      All\\[-2pt]
      (PCK)
    \end{tabular} 
   & \begin{tabular}{c}
      All\\[-2pt]
      (AUC)
    \end{tabular}  \\
\hline\hline
Martinez~\cite{Martinez2017ASY} & H36M & 49.8 & 42.5 & 31.2 & 42.5 & 17.0 \\
Mehta~\cite{mono-3dhp2017} & H36M & 70.8 & 62.3 & 58.8 & 64.7 & 31.7 \\
Yang~\cite{yang2018} & H36M+MPII & - & - & - & 69.0 & 32.0 \\
Zhou~\cite{Zhou_2017_ICCV} & H36M+MPII & 71.1 & 64.7 & 72.7 & 69.2 & 32.5 \\
Luo~\cite{luo2018orinet} & H36M & 71.3 & 59.4 & 65.7 & 65.6 & 33.2 \\
Ci~\cite{wang2019gcn} & H36M & 74.8 & 70.8 & 77.3 & 74.0 & 36.7 \\
Zhou~\cite{Zhou2019HEMletsPL} & H36M+MPII & 75.6 & 71.3 & \textbf{80.3} & 75.3 & 38.0 \\
\hline
Ours & H36M & \textbf{81.5} & \textbf{81.7} & 75.2 & \textbf{80.1} & \textbf{45.8} \\
\hline
\end{tabular}
}
\end{center}
\caption{Results on the MPI-INF-3DHP test set~\cite{mono-3dhp2017}.}
\label{tbl:mpi-inf-3dhp}
\end{table}

\section{Conclusions}
We present a novel architecture for 2D-to-3D human pose estimation, the Graph Stacked Hourglass Networks (GraphSH). With our unique skeletal pooling and skeletal unpooling scheme, together with the proposed architecture
which has powerful multi-scale and multi-level feature extraction capabilities on graph-structured data, our method achieves accurate 2D-to-3D human pose estimation outperforming the state-of-the-art. 
As future work, we hope to introduce temporal multi-frame features in our architecture for further improvement.

{\small
\bibliographystyle{ieee_fullname}
\bibliography{egbib}
}

\end{document}